\documentclass[runningheads]{llncs}
\PassOptionsToPackage{breaklinks,colorlinks,citecolor=blue,linkcolor=blue,urlcolor=blue}{hyperref}
\usepackage{graphicx}
\usepackage{booktabs}

\usepackage[accsupp]{axessibility}  % Improves PDF readability for those with disabilities.

\usepackage{orcidlink}
\usepackage{listings}
\usepackage{microtype}      % microtypography
\usepackage{xcolor}         % colors
\usepackage{colortbl}
\usepackage{makecell}
\usepackage{amsmath}    
\usepackage{csquotes}
\usepackage{multirow}
\usepackage{hyperref}
\hypersetup{breaklinks=true,colorlinks=true,citecolor=blue,linkcolor=blue,urlcolor=blue}

\begin{document}
\raggedbottom

% ---------------------------------------------------------------
% TODO REVIEW: Replace with your title
\title{HumanOmni-Speaker: Identifying Who said What and When} 

% TODO REVIEW: If the paper title is too long for the running head, you can set
% an abbreviated paper title here. If not, comment out.
\titlerunning{HumanOmni-Speaker}

% TODO FINAL: Replace with your author list. 
% Include the authors' OCRID for the camera-ready version, if at all possible.
\author{Detao Bai\inst{1}\and
Xihan Wei\inst{1} \and
Zhiheng Ma\inst{2,3,4}\thanks{Corresponding author: \email{mazhiheng@suat-sz.edu.cn}.}\orcidlink{0000-0002-0034-2065}}

% TODO FINAL: Replace with an abbreviated list of authors.
\authorrunning{D. Bai et al.}
% First names are abbreviated in the running head.
% If there are more than two authors, 'et al.' is used.

% TODO FINAL: Replace with your institution list.
\institute{Tongyi Lab Alibaba Group 
\and Shenzhen University of Advanced Technology \and Guangdong Provincial Key Laboratory of Computility Microelectronics \and  Shenzhen Institutes of Advanced Technology, Chinese Academy of Sciences\\
\email{\{detao.bdt,xihan.wxh\}@alibaba-inc.com, mazhiheng@suat-sz.edu.cn}
\url{https://github.com/HumanMLLM/HumanOmni-Speaker}}

\maketitle

\begin{abstract}
While Omni-modal Large Language Models have made strides in joint sensory processing, they fundamentally struggle with a cornerstone of human interaction: deciphering complex, multi-person conversational dynamics to accurately answer ``Who said what and when.'' Current models suffer from an ``illusion of competence''—they exploit visual biases in conventional benchmarks to bypass genuine cross-modal alignment, while relying on sparse, low-frame-rate visual sampling that destroys crucial high-frequency dynamics like lip movements. To address this limitation, we introduce Visual-Registered Speaker Diarization and Recognition (VR-SDR) and the HumanOmni-Speaker Benchmark. By strictly eliminating visual shortcuts, this rigorous paradigm demands true end-to-end spatio-temporal identity binding using only natural language queries. To overcome the underlying architectural perception gap, we propose HumanOmni-Speaker, powered by a Visual Delta Encoder. By sampling raw video at 25 fps and explicitly compressing inter-frame motion residuals into just 6 tokens per frame, it captures fine-grained visemes and speaker trajectories without triggering a catastrophic token explosion. Ultimately, HumanOmni-Speaker demonstrates strong multimodal synergy, natively enabling end-to-end lip-reading and high-precision spatial localization without intrusive cropping, and achieving superior performance across a wide spectrum of speaker-centric tasks.
% https://github.com/HumanMLLM/HumanOmni-Speaker

\keywords{Multimodal Large Language Models \and Audio-Visual Speaker Diarization \and Audio-Visual Speech Recognition \and Spatio-Temporal Alignment}
\end{abstract}

\section{Introduction}
\label{sec:intro}

\begin{figure}[htbp]
\centering
\includegraphics[width=0.88 \linewidth]{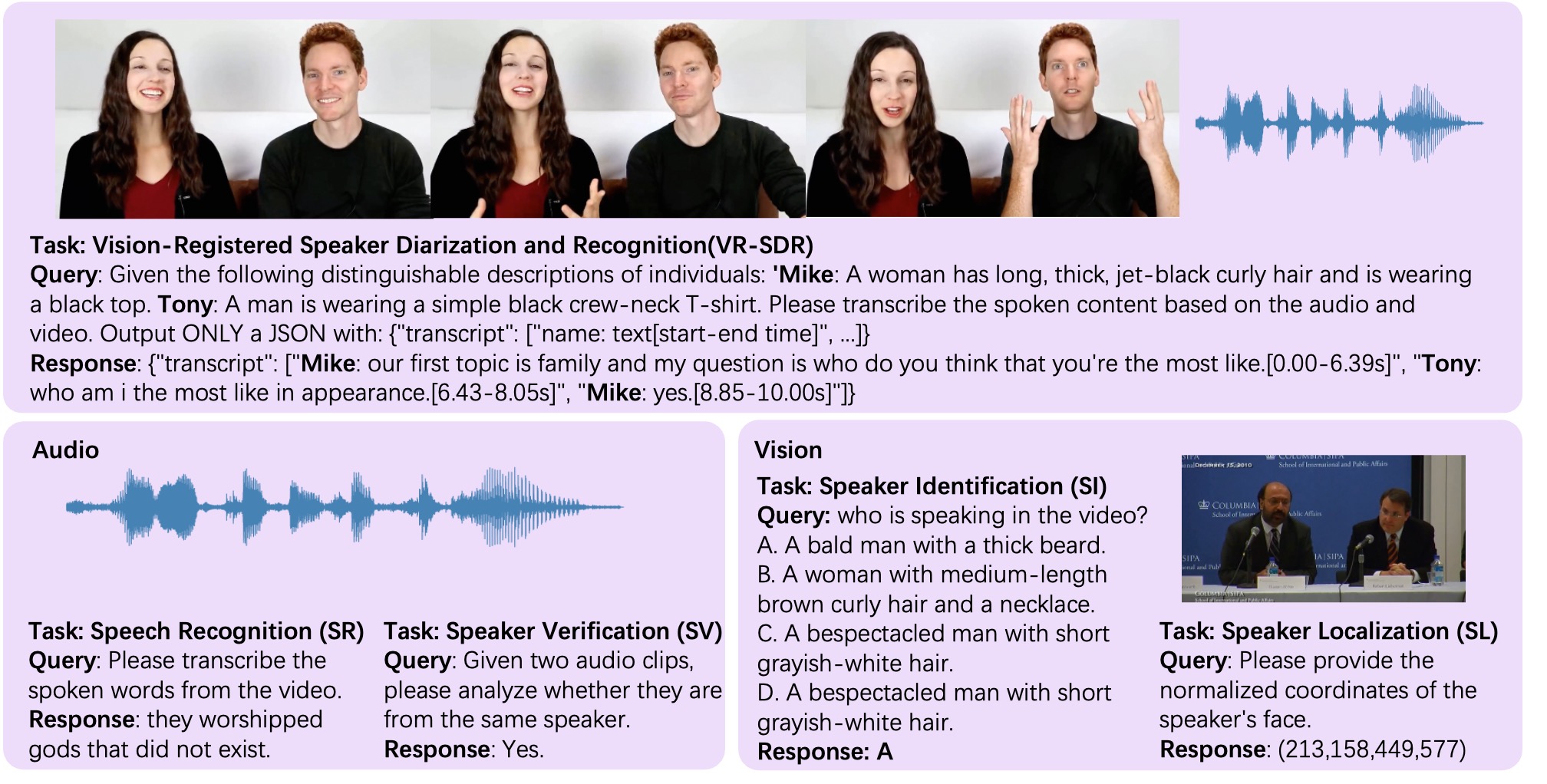}
\vspace{-0.2cm}
\caption{\label{fig:sample} Speaker-Centric examples in HumanOmni-Speaker benchmark. (top) Visual-Registered Speaker Diarization and Recognition (VR-SDR); (bottom) Four atomic subtasks: Speech Recognition (SR), Speaker Verification (SV), Speaker Identification (SI), and Speaker Localization (SL). Our model unifies visual, audio, and text cues in a single framework.}

\end{figure}

The rapid evolution of Large Language Models into Omni-modal architectures—exemplified by Gemini3~\cite{Gemini3}, Qwen3-Omni~\cite{Qwen3Omni}, Qwen2.5-Omni~\cite{Qwen25Omni}, LLaMA-Omni~\cite{LLaMAOmni}, OLA~\cite{Ola}, and VITA~\cite{VITA-1.5}—has enabled the joint processing of visual, audio, and textual signals within a unified semantic space. Yet, as these models transition from isolated inputs to complex real-world environments, seamless human-centric interaction emerges as a critical frontier. At the core of this challenge is the ability to decipher multi-person conversational dynamics by accurately answering: \enquote{\textit{Who said what and when}}. The significance of this capability is twofold. Practically, it is the cornerstone for deploying autonomous embodied agents~\cite{EmbodiedAIAgents}, meeting diarization systems\cite{SpeakerDiarizationReview}  and wearable AI assistants\cite{EmbodiedAIAgents,Ego4D}. Scientifically, it serves as the ultimate stress test for multimodal synergy. It forces models to transcend static semantic matching (e.g., image captioning~\cite{vinyals2015tellneuralimagecaption}) and achieve precise cross-modal spatio-temporal alignment—dynamically binding visual identities, continuous acoustic trajectories, and text under multi-speaker interference. While existing models excel at describing static frames~\cite{Qwen3VL,Ola} or transcribing clear speech~\cite{qwen-audio,qwen2-audio}, they severely struggle in these highly dynamic interactive scenarios.

Why has this fundamental weakness remained largely unaddressed? The answer lies in the deeply flawed evaluation paradigms currently in use. Existing speaker-related tasks\cite{AVA-ASD,Librispeech,Columbia_dataset,SpeakerLM} and Omni benchmarks~\cite{OmniBench,StreamingBench,OmniMMI} isolate the \enquote{\textit{Who}}, \enquote{\textit{When}}, and \enquote{\textit{What}} into siloed, unimodal atomic tasks. More importantly, they inadvertently allow models to bypass true cross-modal understanding through \enquote{shortcut learning} driven by strong visual biases~\cite{visionbiasvgg}. For instance, a model might correctly identify a speaker simply because they are the central subject in a frame or holding a microphone. By relying on such single-modal visual cues, models achieve high scores without ever aligning the audio stream with subtle, continuous facial movements. This pseudo-multimodal alignment creates an illusion of competence, effectively masking the models' profound inability to perform genuine spatio-temporal identity binding.

 % Benchmark\hspace{0pt}---\hspace{0pt}a
% shatter this illusion  修改为 address this limitation
To address this limitation, we must establish a rigorous evaluation paradigm that strictly eliminates these shortcuts. Therefore, we introduce \textbf{Visual-Registered Speaker Diarization and Recognition (VR-SDR)}, an uncompromising task that forces models to output structured records of identities, timestamps, and transcribed content based solely on audio-visual streams and natural language identity descriptions (as illustrated in Fig.~\ref{fig:sample}). When existing Omni-models are subjected to this stringent standard, their true underlying bottleneck is immediately exposed: a severe architectural perception gap. Current models overwhelmingly rely on sparse visual sampling (typically 1-2 fps)—akin to processing only the static \textbf{I-frames} in video compression while discarding the dynamic \textbf{P-frames and B-frames}. This fundamentally destroys the high-frequency temporal dynamics, such as visemes and micro-expressions, essential for tracking speakers. Furthermore, simply increasing the frame rate of standard Vision Transformers (ViTs) to recover this data is catastrophic, as it triggers a quadratic token explosion and drowns subtle motion signals in redundant background noise.

To systematically address both the evaluation illusion and the architectural bottleneck, we propose \textbf{HumanOmni-Speaker}, a speaker-centric unified Omni-model. The main contributions of this study are reflected in the following three aspects:

\begin{enumerate}
    \item \textbf{Redefining evaluation paradigms: bridging the omni-modal capability-assessment gap via the VR-SDR task.} We construct the HumanOmni-Speaker Benchmark, integrating VR-SDR with atomic diagnostic subtasks. By categorizing samples into Easy and Hard levels—where the Hard set explicitly removes visual biases—we rigorously evaluate cross-modal identity consistency, temporal role stability, and comprehensive multimodal understanding without allowing shortcut learning.
    \item \textbf{Architectural Optimization: introducing Visual Delta Encoder to address the shortcomings in high-frequency dynamic perception.} Through the Structured Visual Tokenizer (SVT), the Visual Delta Encoder compresses inter-frame motion residuals into just 6 tokens per frame. This highly condensed representation enables the model to support high-frequency video sampling at 25 fps, allowing it to explicitly track speakers and extract fine-grained visual phonemes (Visemes) without the catastrophic token overhead.
    \item \textbf{HumanOmni-Speaker System: an Omni model tailored for speaker-centric tasks.} Experiments demonstrate that the proposed model significantly outperforms existing open-source models on our benchmark and competes with closed-source models like Gemini3-Pro, effectively addressing the shortcomings in fine-grained motion modeling and cross-modal spatio-temporal alignment. Notably, HumanOmni-Speaker is the first Omni model capable of end-to-end lip-reading and high-precision speaker localization directly from raw video without intrusive lip-cropping pre-processing. We will release the benchmark dataset, code, and model checkpoints upon acceptance.
\end{enumerate}

\section{Related Work}
\label{sec:related_work}

\subsection{Speaker-Centric Tasks and Omni-Benchmarks}

Traditional speaker research decomposes complex interactions into isolated, unimodal subtasks like Automatic Speech Recognition~\cite{Librispeech}, Speaker Diarization~\cite{SpeakerLM}, and Active Speaker Detection~\cite{AVA-ASD}. This \enquote{siloed} paradigm severs the \enquote{Who}, \enquote{When}, and \enquote{What}, masking models' inability to perform genuine cross-modal spatio-temporal binding. Furthermore, existing Omni-modal benchmarks~\cite{VsTaR,OmniBench,StreamingBench,OmniMMI} lack comprehensive speaker-centric evaluation. V-STaR~\cite{VsTaR} and OmniBench~\cite{OmniBench} omit audio-visual speaker tasks, while StreamingBench~\cite{StreamingBench} and OmniMMI~\cite{OmniMMI} are susceptible to visual biases, allowing models to exploit single-modal shortcuts rather than exhibiting true multimodal synergy. To strictly evaluate end-to-end audio-visual alignment and eliminate this illusion of competence, we introduce the VR-SDR task and the comprehensive HumanOmni-Speaker Benchmark.

\subsection{Omni-Modal Large Language Models}

Recent Omni-models—including Gemini3~\cite{Gemini3}, Qwen3-Omni~\cite{Qwen3Omni}, Qwen2.5-Omni~\cite{Qwen25Omni}, LLaMA-Omni~\cite{LLaMAOmni}, OLA~\cite{Ola}, and VITA~\cite{VITA-1.5}—enable end-to-end multimodal processing. However, they typically rely on sparse visual sampling (1-2 fps), capturing static spatial semantics but creating a severe \textit{architectural perception gap} that discards high-frequency dynamics like visemes. While recent work~\cite{fp16llm} increases the frame rate to 16 fps with post-hoc token compression, its underlying static feature extractor remains fundamentally misaligned for continuous motion. Naively increasing sampling rates risks a quadratic \textit{token explosion} and drowns subtle temporal signals in redundant spatial noise. To overcome this, our Visual Delta Encoder operates at 25 fps to explicitly encode inter-frame motion residuals, capturing fine-grained dynamics with minimal token overhead to bridge the audio-visual spatio-temporal gap.

\section{HumanOmni-Speaker Benchmark}

To systematically assess the comprehensive capabilities of Omni-models in complex speaker-centric scenarios, we construct the HumanOmni-Speaker  Benchmark\hspace{0pt}---\hspace{0pt}a hierarchically structured evaluation framework.

At its core is the holistic \textbf{VR-SDR} task, designed to rigorously test the unified understanding of \enquote{Who said what and when} across all modalities. To complement this, we incorporate four atomic subtasks: Speech Recognition (SR), Speaker Verification (SV), Speaker Localization (SL), and Speaker Identification (SI), as depicted in Fig.~\ref{fig:sample}. The inclusion of these atomic tasks serves a critical \textit{diagnostic} purpose: when a model fails the complex VR-SDR task, this hierarchical structure allows us to precisely isolate the underlying bottleneck---determining whether the failure stems from fundamental acoustic perception, spatial visual tracking, or the final cross-modal fusion.

\begin{figure}[htbp]
\centering
\includegraphics[width=0.88 \linewidth]{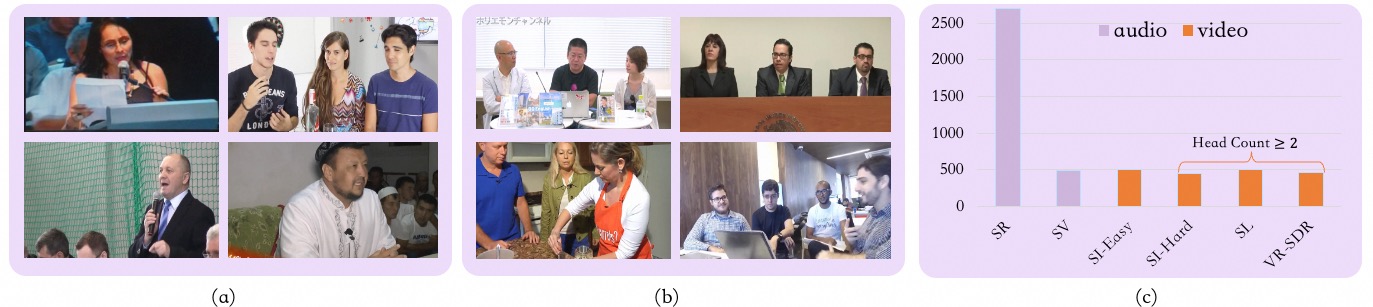}
\vspace{-0.2cm}
\caption{\label{fig:vision_bias}Humanomni-Speaker benchmark sample characteristics and  statistics.(a) Samples with strong visual biases which provide \enquote{recognition shortcuts}. (b) Samples without visual biases that require both acoustic and visual information for speaker identification. (c) Benchmark sample statistics overview.}
\vspace{-0.7cm}
\end{figure}

Beyond the task architecture, the integrity of the evaluation data is equally critical. Existing speaker evaluation datasets are frequently compromised by strong visual biases---such as extreme close-ups, handheld microphones, or single-person compositions (see Fig.~\ref{fig:vision_bias}a). These biases inadvertently provide \enquote{recognition shortcuts,} enabling models to succeed using static visual heuristics rather than genuine audio-visual alignment. To strictly eliminate this shortcut learning in our visual-dependent tasks, we implemented a rigorous \textit{manual filtering} process to categorize the multimodal evaluation into \textbf{Easy} and \textbf{Hard} levels. For the Hard set, human annotators meticulously removed samples containing obvious visual clues. Consequently, the Hard set exclusively comprises complex multi-person scenes and non-subject compositions (Fig.~\ref{fig:vision_bias}b) where speaker identity cannot be reliably deduced from any single static frame. By integrating diagnostic subtasks and enforcing manually curated difficulty levels, the HumanOmni-Speaker Benchmark guarantees an uncompromising, objective evaluation of true cross-modal spatio-temporal capabilities. Detailed task divisions are summarized in Table~\ref{tab:speaker_benchmarks}.

\subsection{Holistic Evaluation: Visual-Registered Speaker Diarization and Recognition (VR-SDR)}

To comprehensively evaluate Omni-models, we propose a novel paradigm for resolving \enquote{\textit{Who said what and when}}---Visual-Registered Speaker Diarization and Recognition (VR-SDR). 

\textbf{Task Definition}: Given an audio-visual input of a multi-person conversation and a set of visual identity registration queries in natural language (e.g., \textit{Alice: A woman with long curly hair; Bob: A man in a black T-shirt}), the model must achieve end-to-end identity binding based solely on visual descriptions. It is required to output structured records containing identity labels, timestamps, and transcribed content (e.g., \textit{Alice: our first topic is family... [0.00-6.39s]; Bob: who am i the most like... [6.43-8.05s]}). 

Unlike traditional audio-only SDR systems~\cite{SpeakerLM} that rely on pre-extracted speaker embeddings for registration, VR-SDR requires only visual semantic descriptions to complete identity binding, completely eliminating the need for prior acoustic enrollment. This task tightly integrates the speakers' visual and audio streams with textual identity cues, aligning perfectly with the language-centric, multimodal interaction logic of Omni-models. 

\textbf{Evaluation Metrics}: We employ \textit{Identity-Fixed} metrics to rigorously evaluate \enquote{\textit{Who said what}} and \enquote{\textit{Who said when}}.

\textbf{Identity-Fixed SA-WER (Who said what):} Unlike the traditional concatenated minimum-permutation word error rate (cpWER), which allows for optimal label re-alignment, we employ Speaker-Attributed Word Error Rate (SA-WER). This is a strictly harsher metric requiring the model's output to precisely correspond to the vision-registered ID. A word is counted as correct only if both the text and the assigned Speaker ID are accurate. This directly measures the model's capacity to bind visual descriptions with acoustic transcriptions. Formally, given a set of registered speakers $\mathcal{S}$, it is calculated as:
\begin{equation}
    \text{SA-WER} = \frac{\sum_{s \in \mathcal{S}} \text{EditDistance}(\text{Ref}_s, \text{Hyp}_s)}{\sum_{s \in \mathcal{S}} |\text{Ref}_s|}
\end{equation}
where $\text{Ref}_s$ is the reference transcript for speaker $s$, and $\text{Hyp}_s$ is the hypothesized transcript that the model attributes to speaker $s$. 

\textbf{Identity-Fixed IER (Who said when):} The Identification Error Rate (IER) evaluates the model's capability in Identity-Fixed speaker diarization, simultaneously assessing identity binding accuracy and temporal alignment precision. Compared to the traditional Speaker Diarization Error Rate (DER), which allows for label permutation to minimize penalties, IER enforces an absolute mapping between the model output and the vision-registered identities $\mathcal{S}$. Formally, it is defined as:
\begin{equation}
    \text{IER} = \frac{\sum_{s \in \mathcal{S}} (\text{Miss}_s + \text{FA}_s + \text{Conf}_s)}{\sum_{s \in \mathcal{S}} \text{Dur}_s}
\end{equation}
where $\text{Miss}_s$, $\text{FA}_s$, and $\text{Conf}_s$ represent the durations of missed speech, false alarms, and speaker confusion attributed to speaker $s$, respectively, and $\text{Dur}_s$ denotes the total ground-truth duration for speaker $s$.

\subsection{Diagnostic Evaluation and Construction Details}

To systematically analyze performance bottlenecks in the holistic VR-SDR task, we designed a progressive diagnostic chain of four atomic subtasks—ranging from basic acoustic competence to full cross-modal identity binding. To construct this benchmark, we implemented a rigorous annotation pipeline. After filtering source datasets~\cite{AVSpeech,VoxMM,Librispeech,Vox2,Columbia_dataset}, we utilized Qwen-VL~\cite{Qwen3VL} to generate foundational appearance descriptions and QA pairs. A team of 10 English-proficient annotators (TEM-8 certified or equivalent) and three senior inspectors then performed meticulous calibration to guarantee high-fidelity annotations (Fleiss' $\kappa = 0.82$ on ${\sim}$1.5K samples; 18\% disagreements resolved by inspector adjudication). For the Easy/Hard partitioning, a sample is labeled \textit{Easy} if it exhibits any salient visual shortcut (close-up, visible microphone, single-person composition, or centrally lit speaker), and \textit{Hard} otherwise. The resulting tasks are structured as follows:

\begin{itemize}
    \item \textbf{Speech Recognition (SR):} Underpins the \enquote{What} dimension by evaluating basic acoustic content transcription. Evaluated using LibriSpeech-Test-Clean~\cite{Librispeech}.
    \item \textbf{Speaker Verification (SV):} Probes acoustic identity discrimination (the \enquote{Who} dimension) without visual cues. Built from the VoxCeleb2~\cite{Vox2} test set, it comprises 486 binary QA pairs (1:1 positive-negative ratio).
    \item \textbf{Speaker Localization (SL):} Evaluates audio-visual spatial perception by requiring end-to-end localization of the active speaker. Extracted from Columbia-ASD~\cite{Columbia_dataset} (500 multi-person clips), it forces models to directly output face coordinates from raw video, bypassing traditional cascaded \enquote{detection--cropping--classification} pipelines.
    \item \textbf{Speaker Identification (SI):} The atomic task closest to VR-SDR. It tests cross-modal identity binding by requiring the model to match a natural language visual description to the active speaker under visual interference. Sourced from AVSpeech~\cite{AVSpeech} using 4-option multiple-choice questions.
    \item \textbf{VR-SDR (Holistic):} The ultimate stress test. Sourced from VoxMM~\cite{VoxMM}, it comprises $\sim$464 multi-person conversation segments, with descriptions iteratively refined via human-LLM collaboration to ensure unique identity binding.
\end{itemize}

Within this framework, SR and SV serve as audio-only baselines, whereas VR-SDR, SL, and SI demand audio-visual processing of multi-person scenarios. To ensure visually dependent tasks are not confounded by spurious heuristics, we partitioned the SL and SI samples into Easy and Hard subsets. For the Hard sets, we manually filtered out samples exhibiting significant visual biases, strictly mandating genuine temporal audio-visual alignment. Evaluation metrics and sample distributions are summarized in Table~\ref{tab:atomic_evaluation} and Fig.~\ref{fig:vision_bias}(c), respectively.

\begin{table}[htbp]
\vspace{-0.2cm} 
\caption{The Hierarchical Evaluation Matrix and Benchmark Statistics}. 
\label{tab:atomic_evaluation}
\vspace{-0.2cm} 
\centering
\small
\resizebox{0.99\textwidth}{!}{
\begin{tabular}{lcccc@{}cccc}
\toprule
  & \multicolumn{4}{c}{\textbf{Evaluation Framework}} & \multicolumn{4}{c}{\textbf{Benchmark Statistics}} \\
  \cmidrule(lr){2-5} \cmidrule(lr){6-9}
  \textbf{Level} &\textbf{Speaker-Task}&  \textbf{Modality}&\textbf{Core Competency} & \textbf{Metric}  &\textbf{Samples}  & \textbf{Duration} & \textbf{Avg Pers.}&\textbf{Avg Spk.} \\ \midrule
  \textbf{Holistic} &\makecell{Vision-Registered Speaker\\ Diarization Recognition (VR-SDR)}&  \textbf{A}-\textbf{V}-\textbf{T}&\makecell{Omni-modal comprehension \\ across audio, text, and video } & \makecell{SA-WER(\%) $\downarrow$ \\IER  (\%) $\downarrow$} & 464 & 1.2h & 3.4&2.1 \\\midrule
  \multirow{4}{*}{\textbf{Atomic}} &Speech Recognition (SR)& \textbf{A}& Acoustic Content Transcription&WER (\%) $\downarrow$  & 2620 & 5.4h & 1 &1 \\
  &Speaker Verification (SV)&  \textbf{A}&Acoustic Identity Discrimination& Error (\%) $\downarrow$  & 486 & 1.0h& 1.5 &1.5\\
  &Speaker Localization (SL)&  A-\textbf{V}&Audio-Visual Spatial Perception& Miss Rate (\%)$\downarrow$ & 500 & 0.62h& 2.6&1 \\
  &Speaker Identification (SI)&  A-\textbf{V}-\textbf{T}&Audio-Visual Identity Binding& Error (\%) $\downarrow$  & 942& 1.3h& 3.2&1 \\ \bottomrule
\end{tabular}
}
\end{table}

% \begin{table}[htbp]
% \vspace{-0.2cm} 
% \caption{The Hierarchical Evaluation Matrix. }
% \label{tab:atomic_evaluation}
% \vspace{-0.2cm} 
% \centering
% \small
% \resizebox{0.9\textwidth}{!}{
% \begin{tabular}{lcccc@{}}
% \toprule
%   \textbf{Level} &\textbf{Speaker-Task}&  \textbf{Modality}&\textbf{Core Competency} & \textbf{Metric} \\ \midrule
%   \textbf{Holistic} &\makecell{Vision-Registered Speaker\\ Diarization Recognition (VR-SDR)}&  \textbf{A}-\textbf{V}-\textbf{T}&\makecell{Omni-modal comprehension \\ across audio, text, and video } & \makecell{SA-WER(\%) $\downarrow$ \\IER  (\%) $\downarrow$}\\\midrule
%   \multirow{4}{*}{\textbf{Atomic}} &Speech Recognition (SR)& \textbf{A}& Acoustic Content Transcription&WER (\%) $\downarrow$ \\
%   &Speaker Verification (SV)&  \textbf{A}&Acoustic Identity Discrimination& Error (\%) $\downarrow$ \\
%   &Speaker Localization (SL)&  A-\textbf{V}&Audio-Visual Spatial Perception& Miss Rate (\%)$\downarrow$\\
%   &Speaker Identification (SI)&  A-\textbf{V}-\textbf{T}&Audio-Visual Identity Binding& Error (\%) $\downarrow$ \\ \bottomrule
% \end{tabular}
% }
% \end{table}

\section{Architecture}

\subsection{Overview}

\begin{figure}[htbp]
\centering
\includegraphics[width=0.95 \linewidth]{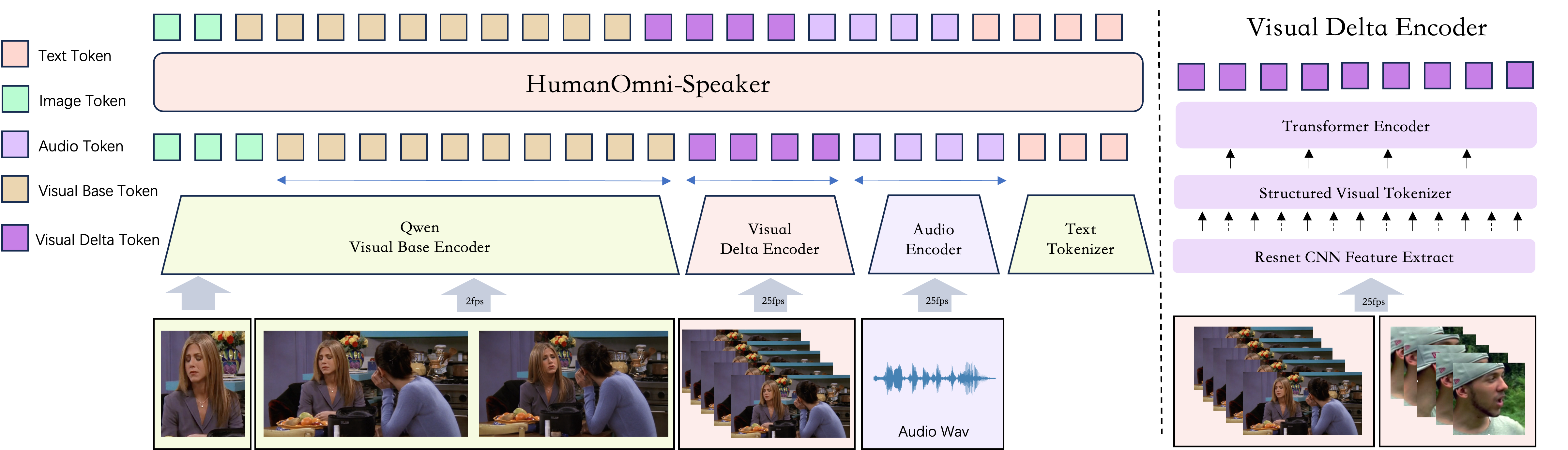}
\vspace{-0.2cm}
\caption{\label{fig:archmodel} Overview of HumanOmni-Speaker architecture for human-centric speaking scenarios. It integrates text, audio, and visual inputs through Text Tokenizer, Audio Encoder, Visual Base Encoder (1-2 fps), and Visual Delta Encoder (25 fps). The Visual Delta Encoder (right) employs a ResNet-18 backbone, Spatio-Temporal Vision Transformer (SVT), and Transformer encoder, producing only 6 structured tokens per frame. All modality-specific tokens are aligned in a shared LLM decoder, enabling the model to reason about \enquote{Who}, \enquote{When} and \enquote{What} in complex interactions. }
\vspace{-0.7cm}
\end{figure}

HumanOmni-Speaker is an Omni model specifically built for human-centric speaking scenarios. As shown in Fig.\ref{fig:archmodel}, the architecture serializes visual spatial, visual temporal, audio, and text features into tokens via the Visual Base Encoder, Visual Delta Encoder, Audio Encoder, and Text Tokenizer. These tokens are aligned and fed into an LLM for decoding. Multimodal information is fused through a cross-attention mechanism to achieve a deep understanding of \enquote{Who}, \enquote{When} and \enquote{What} in complex interactions. 

Regarding visual representation, HumanOmni-Speaker utilizes a dual-stream path to explicitly deconstruct spatial semantics and temporal dynamics. The first stream, the \textbf{Visual Base Encoder}, shares its backbone with Qwen2.5-Omni and extracts stable identity features and environmental context through low-frequency sampling (1-2 fps). The second stream is the \textbf{Visual Delta Encoder}, our core innovation, explicitly models inter-frame residuals via 25 fps high-frequency sampling, capturing temporal details and high-frequency features that are ignored by Visual Base Encoder due to sparse sampling. This design ensures high-precision perception of speaker-centric scenes while maintaining a low computational load through functional specialization. The audio and text encoders remain consistent with Qwen2.5-Omni. 

\subsection{Visual Delta Encoder}
Unlike traditional image encoders, the Visual Delta Encoder explicitly extracts fine-grained, inter-frame temporal dynamics. As depicted in Fig.~\ref{fig:archmodel} (right), it employs a three-stage architecture: \textbf{Local Feature Perception} utilizes a lightweight ResNet-18 backbone to balance computational efficiency with local motion capture at 25 fps. \textbf{Structured Visual Tokenizer (SVT)} applies hierarchical spatial ($7\times7$) and large-receptive-field temporal ($k=63$) convolutions to compress dense CNN features into just 6 structured tokens per frame. This mitigates the sequence load of high-frequency sampling while preserving high-fidelity lip visemes and motion trajectories. \textbf{Global Context Encoding} leverages a Transformer Encoder to process these tokens across frames, integrating discrete temporal increments into coherent behavioral semantics.

\begin{figure}[htbp]
\centering
\includegraphics[width=0.88 \linewidth]{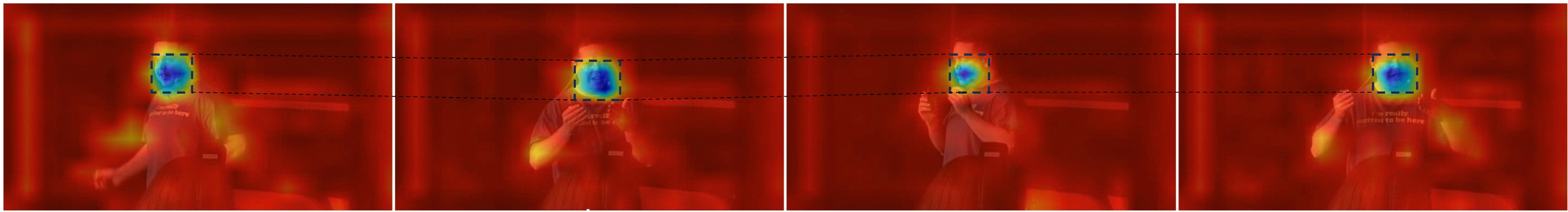}
\vspace{-0.2cm}
\caption{\label{fig:gradcam}The attention maps generated by Grad-CAM show that Visual Delta Encoder successfully localizes and tracks the speaker's mouth. }
\vspace{-1cm}
\end{figure}

This architecture empowers the Visual Delta Encoder with two critical capabilities:

\begin{itemize}
    \item \textbf{Spatio-Temporal Speaker Tracking:} Empowered by the 25 fps sampling rate, the encoder extracts highly continuous spatial trajectories. Grad-CAM visualizations (Fig.~\ref{fig:gradcam}) confirm that its shallow CNN layers consistently lock onto the speaker's face and mouth, providing the underlying motion primitives required for VR-SDR.
    \item \textbf{Fine-Grained Viseme Modeling:} The module directly extracts visual phonemes (visemes) from raw video streams, bypassing the need for intrusive face alignment or lip-cropping. As evidenced in Table~\ref{tab:vsr_avsr}, this inter-frame residual representation successfully overcomes the historical limitations of Omni-models in Visual Speech Recognition (VSR).
\end{itemize}

Through the synergy of trajectory tracking and viseme modeling, the Visual Delta Encoder serves as a vital \enquote{high-frequency cross-modal bridge.} By outputting dense spatio-temporal tokens, it seamlessly spans the gap between high-frequency audio streams and sparse, high-level visual semantics. Ultimately, it transcends basic feature extraction, directly enabling the Omni-model to spatially and temporally align multi-speaker dynamics and reliably resolve \enquote{Who said what and when.}

\section{Training}

\begin{figure}[htbp]
\centering
\includegraphics[width=0.88 \linewidth]{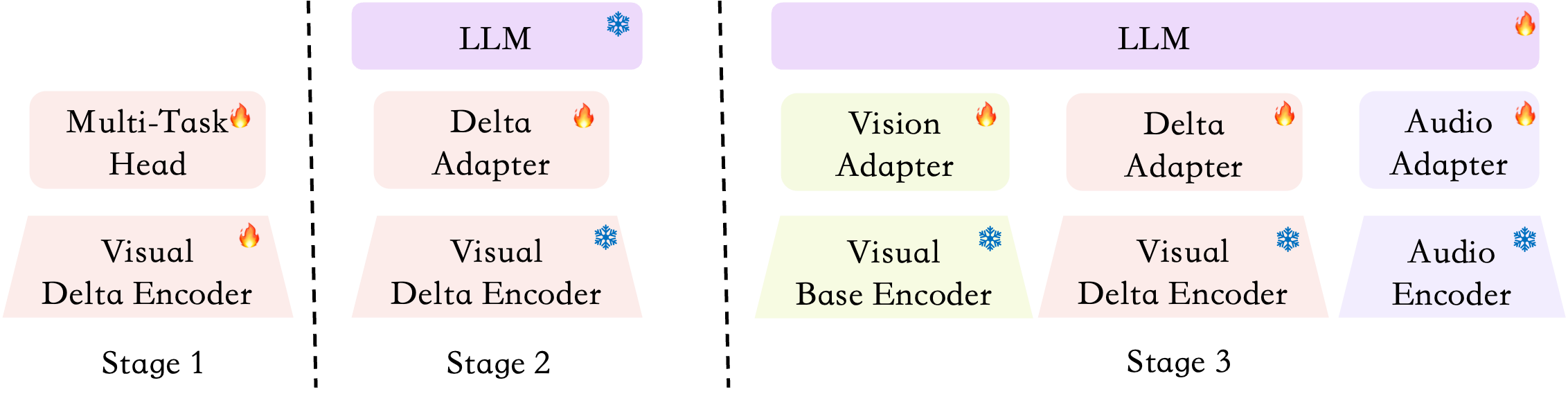}
\vspace{-0.2cm}
\caption{\label{fig:trainpipeline} 
The Progressive Training Pipeline of HumanOmni-Speaker.
}
\end{figure}

We employ a progressive three-stage training paradigm to systematically equip the model with high-frequency dynamic perception and cross-modal alignment capabilities (Fig.~\ref{fig:trainpipeline}). 

\textbf{Stage 1: Visual Delta Encoder Pre-training}. To master high-frequency motion primitives, we pre-train the Visual Delta Encoder on a curated large-scale speaker-centric dataset aggregating AVSpeech, LRS2, LRS3, VoxCeleb2, and AVA-ASD~\cite{AVSpeech,LRS2,LRS3,Vox2,AVA-ASD}. We formulate this as a multi-task supervised learning framework. Using a shared encoder backbone with task-specific heads, we simultaneously optimize for spatial tracking (via face center-point regression) and fine-grained lip-reading (using VSR and AVSR objectives from CoGenAV~\cite{CoGenAV}). This simultaneous optimization endows the encoder with the fundamental capability to continuously track spatial displacements and extract robust viseme features in the temporal domain.

\textbf{Stage 2: Modality Alignment}. Next, we align the visual delta features with the latent semantic space of the LLM. Using the pre-training dataset, we freeze both the Visual Delta Encoder and the LLM decoder, optimizing only a lightweight MLP projector. Crucially, we enforce \textit{modality isolation} during this phase: the original Visual Base Encoder and Audio Encoder from Qwen2.5-Omni are bypassed, and the LLM receives features exclusively from the Visual Delta Encoder. This strict isolation forces the LLM to effectively translate high-fidelity viseme details and motion trajectories into interpretable behavioral semantics, preventing it from relying on representational shortcuts from established modalities.

\textbf{Stage 3: End-to-End Joint Fine-Tuning}. Finally, we construct a comprehensive dataset of approximately 1.5 million audio-visual instructions tailored for speaker-centric scenarios, encompassing ASR, VSR, Speaker Localization, Verification, Identification, and VR-SDR. We standardize all unimodal and multimodal inputs (raw video frames, continuous audio waveforms, and text prompts) into a unified instruction format. 

During this joint optimization phase, all structural encoders remain strictly frozen while their respective MLP projectors are updated. For the LLM decoder, we employ a hybrid parameter-efficient tuning strategy: the initial layers undergo full fine-tuning to promote deep, early-stage cross-modal fusion, while the deeper layers are adapted via LoRA. This strategy maximizes multimodal spatio-temporal synergy while preserving the foundational generation quality of the underlying LLM.

\section{Experiment}
\subsection{HumanOmni-Speaker Benchmark}

Table~\ref{tab:speaker_benchmarks} reports the optimal prompt-tuned performance of various models across the HumanOmni-Speaker Benchmark. 

\textbf{Acoustic Perception: Speech Transcription and Voiceprint Analysis.} While most Omni-models achieve near-ceiling performance in basic Speech Recognition (ASR), the Speaker Verification task exposes a critical flaw in open-source voiceprint extraction. Models like OLA and VITA1.5 perform near random chance, and the Qwen-Omni series exhibits severe distribution bias (over-predicting distinct speakers). In contrast, the closed-source Gemini3-Pro demonstrates robust discriminative capability with a 5.2\% error rate. HumanOmni-Speaker successfully bridges this open-source gap, achieving a 13.2\% error rate and significantly outperforming its base Qwen2.5-Omni-3B model.

\textbf{Visual Perception: High-Precision Speaker Localization and Identity Binding.} HumanOmni-Speaker dominates the Speaker Localization task with a mere 0.8\% error rate, drastically outperforming both Qwen3-Omni (2.8\%) and Gemini3-Pro (12.8\%). This stark contrast underscores a direct correlation between spatial accuracy and visual sampling frequency. While Gemini and Qwen-Omni rely on sparse 1--2 fps sampling, our dual-rate architecture (incorporating the 25 fps Visual Delta Encoder) successfully captures the high-frequency dynamics necessary for precise speaker tracking. 

Furthermore, the Speaker Identification task reveals a profound performance gap between the Easy and Hard sets. In the Easy set, all models exhibit low error rates (mostly $<10\%$), indicating a baseline ability to associate captions with clear visual subjects. However, in the Hard set—which features extreme multi-speaker interference and removes visual shortcuts—all baseline models suffer catastrophic performance degradation. This confirms that existing Omni-models heavily rely on static visual heuristics, whereas robust cross-modal identity binding in dynamic environments remains a critical frontier that our architecture begins to address.

\begin{table}[tbp]
\vspace{-0.5em} 
\caption{Results on HumanOmni-Speaker Benchmarks. The best results are highlighted. }
\label{tab:speaker_benchmarks}
\centering
\small
\resizebox{1.0\textwidth}{!}{ 
\begin{tabular}{lcccccccccc}
\toprule
\textbf{Method} &\makecell{\textbf{Model} \\ \textbf{Size } }  &\makecell{\textbf{Speech} \\ \textbf{Recognition} }  &\makecell{\textbf{Speaker} \\ \textbf{Verification} }  &\makecell{\textbf{Speaker} \\ \textbf{Localization} }  & \multicolumn{2}{c}{\makecell{\textbf{Speaker} \\ \textbf{Identification} }}&  \makecell{\textbf{Atomic} \\ \textbf{AVG} }  &\makecell{\textbf{VR-SDR} \\ \textbf{What} }  &\makecell{\textbf{VR-SDR} \\ \textbf{When} }  & \makecell{\textbf{Holistic} \\ \textbf{AVG} }  \\
 & & easy($\downarrow$)& easy($\downarrow$)&easy($\downarrow$)& easy($\downarrow$)& hard($\downarrow$)&  ($\downarrow$)&hard($\downarrow$)&hard($\downarrow$)&($\downarrow$)\\ 
\midrule
\rowcolor{gray!20}\multicolumn{11}{c}{Closed-source Omni Model}\\
 Gemini3-Pro& -& 1.39&\textbf{5.2}&12.8&  5.5&30.5&  11.1&\textbf{36.6}&36.3&\textbf{36.5}\\
 Qwen3-Omni-flash& -& \textbf{1.22}&43.9&2.8&  3.6&43.5&  19.0&82.9&47.2&65.0\\ 
\rowcolor{gray!20} \multicolumn{11}{c}{Open-source Omni Model}\\
 OLA~\cite{Ola}& 7B& 1.9& 51.1&20.6& 12.4& 63.2&  29.8&95.4& 56.85&76.1\\ 
VITA1.5~\cite{VITA-1.5}&7B& 3.4&51.4&20.2&  10.3&56.6&   28.4&93.6&54.40& 74\\
 Qwen2.5-Omni~\cite{Qwen25Omni}& 3B& 2.2& 44.2& 7.4& 6.6& 51.5&  22.4&84.6& 50.9&67.8\\
 Qwen2.5-Omni~\cite{Qwen25Omni}& 7B& 1.8&37.1&7.4& 4.0& 54.5&  20.9&83.6&49.4&66.5\\
 \rowcolor{gray!20}\multicolumn{11}{c}{HumanOmni-Speaker Model}\\
 Qwen2.5-Omni-SFT& 3B & 2.0&13.4&3.0& 1.7& 33.2&   10.7&52.1&31.5&41.8\\ 
\rowcolor{gray!20} 
\textbf{HumanOmni-Speaker} &3B&1.9&13.2&\textbf{0.8}& \textbf{1.0}&\textbf{21.1}&  \textbf{7.6}&47.1&\textbf{28.5}& 37.8\\ 
\bottomrule
\end{tabular}
} 
\end{table}

\textbf{Multimodal Understanding: Vision-Registered Speaker Diarization and Recognition.} As a comprehensive stress test, the VR-SDR task demands the seamless integration of textual queries, visual streams, and audio signals to jointly resolve \enquote{Who}, \enquote{When}, and \enquote{What}. Our findings reveal that most open-source Omni-models---such as the Qwen-Omni series, OLA, and VITA1.5---falter under the strict demands of full-modality fusion. In contrast, Gemini3-Pro establishes a robust closed-source baseline, achieving 36.6\% and 36.3\% on SA-WER and IER, respectively. This profound performance gap underscores that current open-source models largely fail to achieve genuine spatio-temporal binding between visual identities and acoustic dynamics, indicating that true Omni-modal collaborative understanding remains in its infancy. 

% 原标题 \textbf{HumanOmni-Speaker Performance.}
\textbf{Holistic Analysis: HumanOmni-Speaker Performance and Remaining Challenges.} Comparative analysis against a Qwen2.5-Omni-SFT baseline (lacking the Visual Delta Encoder) highlights the critical necessity of our proposed module in deciphering complex, dynamic multi-speaker environments. Integrating the Visual Delta Encoder yielded a dramatic reduction in Speaker Localization error, dropping from 3.0\% to a remarkable 0.8\%. Concurrently, it drove a substantial decrease in the Speaker Identification Hard Set error rate, lowering it from 33.2\% to 21.1\%. Beyond spatial localization, this high-frequency motion encoding catalyzes true full-modality synergy. In the demanding VR-SDR task, the module optimized the \enquote{Who said what} and \enquote{Who said when} metrics from 52.1\% and 31.65\% down to 47.1\% and 28.5\%, respectively. Decomposing the residual 47.1\% SA-WER reveals that missed segments account for ${\sim}$6\%, identity misattribution for ${\sim}$25\% (consistent with SI-Hard error), and transcription errors for the remainder (speaker-blind WER~=~21.0\%)---confirming that cross-modal identity binding is the primary bottleneck. Moreover, as speaker count grows from 2 to 3, both SA-WER and IER degrade substantially; Gemini3-Pro exhibits the same trend, indicating that multi-party overlapping speech remains a systematic challenge for current Omni architectures regardless of model scale.These gains conclusively validate the Visual Delta Encoder as an essential mechanism for enforcing robust spatio-temporal alignment and achieving the deep cross-modal binding required to definitively resolve \enquote{Who said what and when}.

% \textcolor{red}{\textbf{Error Decomposition.} To pinpoint the remaining bottleneck, we decompose HumanOmni-Speaker's 47.1\% SA-WER into three sources: (i)~\textit{missed segments} (IoU~$<$~0.3 with any ground-truth window) account for ${\sim}$6\%; (ii)~\textit{identity misattribution}---evaluating speaker matching on ground-truth temporal windows yields 74.55\% accuracy, implying ${\sim}$25\% misattribution, consistent with our SI-Hard error of 21.1\%; (iii)~\textit{transcription errors}---when speaker identity is ignored, WER drops to 21.0\%. This decomposition confirms that multi-speaker identity binding, rather than acoustic transcription, constitutes the primary performance bottleneck for current Omni-models on VR-SDR.}

% \textcolor{red}{\noindent\textbf{Scalability with Speaker Count.}
% We further analyze how VR-SDR performance scales with the number of speakers. As the speaker count increases from 2 to 3, SA-WER rises from 46.16\% to 66.11\% and IER from 27.15\% to 44.69\%. Notably, Gemini3-Pro exhibits a similar degradation trend, suggesting that overlapping speech and increased identity interference in multi-party scenarios remain a systematic challenge for current Omni-model architectures regardless of model scale. }

\subsection{Visual and Audio-Visual Speech Recognition}

Due to the fine-grained dynamic representations provided by Visual Delta Encoder, HumanOmni-Speaker excels in visual speech recognition (VSR) and audio-video speech recognition (AVSR). We validated the model's performance on LRS2 and LRS3, and the results are shown in Table  \ref{tab:vsr_avsr}. 

\textbf{Compared to Specific VSR models: No preprocessing required}. Traditional specific VSR models such as Av-HuBERT and Auto-AVSR  rely on complex pre-processing including face alignment and lip region of interest (ROI) cropping.
HumanOmni-Speaker achieves comparable performance to Auto-AVSR on the VSR task of both the LRS3 (33.4\% vs 33.0\%) and LRS2 (27.9\% vs 29.8\%)  without any pre-processing of the raw dataset. 
 
\textbf{Compared to LLM-based AVSR models: Stronger audio-video fusion performance. }Compared to recent LLM-based AVSR models such as Llama-AVSR and Whisper-flamingo, HumanOmni-Speaker delivers exceptional performance in core AVSR metrics. Our model delivers competitive AVSR performance, yielding 0.76\% WER on LRS3 and 1.36\% on LRS2, outperforming other LLM-based models.

\textbf{Compared to Omni models: Filling the Lip-Reading Gap. }Because existing Omni models focus on global visual and audio understanding and lack mechanisms for capturing high-frequency dynamic features, they are often ineffective when faced with VSR and AVSR tasks. HumanOmni-Speaker is the first Omni model to support end-to-end lip-reading natively, filling the gap in fine-grained motion modeling of Omni models. 

\textbf{Comparing the performance of the HumanOmni-Speaker model on ASR and AVSR. }we observe that adding visual information significantly reduces the WER from 3.63\% to 0.76\% on LRS3, and from 3.47\% to 1.36\% on LRS2. This quantitatively demonstrates that lip movement features extracted by the Visual Delta Encoder strongly enhance the audio signal and effectively serve as a semantic bridge across modalities. 

\begin{table}[tbp]
\centering

\begin{minipage}{0.46\textwidth} % 第一个表格的宽度
\centering
\small 
\caption{Results on VSR and AVSR.}
\label{tab:vsr_avsr}
\vspace{-0.2cm}
\resizebox{\textwidth}{!}{ 
\begin{tabular}{llcc}
\toprule
\textbf{Method}  &\textbf{Preprocessing}& \makecell{\textbf{LRS2} \\ \textit{vsr} | \textit{asr} |\textit{avsr}} & \makecell{\textbf{LRS3} \\ \textit{vsr} | \textit{asr} |  \textit{avsr}} \\ 
\midrule
\rowcolor{gray!20} \multicolumn{4}{c}{Specific VSR Model}\\
CTC/Attention~\cite{ctc_attentin} & Lip crop \& align & 63.5|-|7.0 & - \\
AV-HuBERT Base~\cite{av-hubert}~\cite{VatLM} & Lip crop \& align & 31.2|-|- & 34.8|-|- \\ 
AutoAVSR~\cite{Auto-avsr} & Lip crop \& align & 27.9|-|1.5 & 33.0|-|0.9 \\ 
\rowcolor{gray!20} \multicolumn{4}{c}{LLM-base AVSR Model}\\ 
Llama-SMoP~\cite{Llama-SMoP} & Lip crop \& align & - & -|-|0.96 \\ 
Llama-AVSR~\cite{Llama-avsr} & Lip crop \& align & - & 24.0|0.79|0.77 \\ 
Whisper-flamingo~\cite{Whisper-flamingo} & Lip crop \& align & -|-|1.4 & -|-|0.76 \\ 
\rowcolor{gray!20}\multicolumn{4}{c}{Omni Model}\\
OLA~\cite{Ola} & Raw video & -|5.5|- & -|4.7|- \\ 
Qwen2.5-Omni~\cite{Qwen25Omni} & Raw video & -|3.47|- & -|3.63|- \\ 
\rowcolor{gray!20} 
\textbf{HumanOmni-Speaker} & Raw video & 29.8|3.47|\textbf{1.36} & 33.4|3.63|\textbf{0.76} \\
\bottomrule
\end{tabular}
}

\end{minipage}
\hfill
\begin{minipage}{0.50\textwidth} % 第二个表格的宽度
\centering
\caption{Results on ASR.}
\label{tab:audio_benchmarks}
\vspace{-0.2cm}
\resizebox{\textwidth}{!}{ 
\begin{tabular}{llcc}
\toprule
\textbf{Method}  &\textbf{Model Size} & \makecell{\textbf{Librispeech-dev} \\ \textit{dev-clean} | \textit{dev-other}} & \makecell{\textbf{Librispeech-test} \\ \textit{test-clean} | \textit{test-other}} \\ 
\midrule
\rowcolor{gray!20} \multicolumn{4}{c}{Audio Models}\\
 Whisper-small~\cite{whisper}& 0.3B& 4.4|10.1&4.6|10.3\\
SenseVoice-L~\cite{SenseVoice} &  1.6B & - & 2.6|4.3 \\
Qwen-Audio~\cite{qwen-audio} & 7B & 1.8|4.0 & 2.0|4.2 \\
Qwen2-Audio~\cite{qwen2-audio} & 7B & 1.3|3.4 & 1.6|3.6 \\ 
\rowcolor{gray!20} \multicolumn{4}{c}{Omni Models}\\
HumanOmni~\cite{HumanOmni} & 7B & 3.8|7.5 & 3.7|8.0 \\ 
VITA~\cite{VITA-1.5} & 7B & 7.6|16.6 & 8.1|18.4 \\ 
Mini-Omni2~\cite{Mini-Omni2} & 7B & 4.7|9.4 & 4.8|9.8 \\ 
OLA~\cite{Ola} & 7B & 1.9|4.4& 1.9|4.2\\ 
Qwen2.5-Omni~\cite{Qwen25Omni} & 3B & 2.0|4.1 & 2.2|4.5 \\ 
\rowcolor{gray!20} 
\textbf{HumanOmni-Speaker} & 3B & 1.91|4.16 & 1.88|4.56 \\ 
\bottomrule
\end{tabular}
}

\end{minipage}
\end{table}

\subsection{Automatic Speech Recognition}
To evaluate HumanOmni-Speaker on ASR task, we compared its performance with Audio-LLMs and omni models on LibriSpeech. As shown in Table \ref{tab:audio_benchmarks}.

HumanOmni-Speaker (3B) outperforms existing open-source omni models on LibriSpeech, including HumanOmni, VITA, Mini-Omni2, and OLA. It achieves comparable performance  to the base model Qwen2.5-Omni (3B), and even slightly outperforms it on clean subsets, with 1.91\% and 1.88\% WER (vs. 2.0\% and 2.2\%) on dev-clean and test-clean, respectively. Compared to dedicated Audio-LLMs, it also achieves comparable performance with a smaller parameter set.

This indicates that HumanOmni-Speaker retains the strong ASR capabilities of Qwen2.5-Omni. We suggest that our design of the model structure and training method is crucial. In terms of model structure, while incorporating new speaker-aware modules Visual Delta Encoder to enhance visual modeling, it retains the original audio encoder and language decoder backbone. Regarding training methods, the three-stage training strategy allows the model to effectively integrate dynamic speaker-related visual information without interfering with the original speech recognition pathway. 

% 原版的没有小结
\section{Ablation Study}
We conduct a series of ablation experiments to validate the design choices of the Visual Delta Encoder from four perspectives: component effectiveness, end-to-end design advantages, hyperparameter sensitivity, and computational cost.

\noindent\textbf{Component Effectiveness.}
As shown in Table~\ref{tab:ablation_study}, using only the Visual Delta Encoder (without Visual Base Encoder) on the Speaker Localization task achieves a 1.2\% error rate, already significantly better than the base model (3.0\%), indicating that the Visual Delta Encoder alone exhibits strong spatial awareness. Combining both encoders further reduces the error to 0.8\%. On the more complex VR-SDR task, integrating the Visual Delta Encoder reduces SA-WER from 52.1\% to 47.1\% and IER from 31.5\% to 28.5\%.This indicates that the Visual Delta Encoder, by enhancing visual awareness of speaker location, acts as an effective semantic bridge across modalities, aligning high-frequency audio with sparse visual information and enabling deeper binding of \enquote{Who said what and when}.

% First, we evaluate HumanOmni-Speaker using only the proposed Visual Delta Encoder features without Visual Base Encoder features on the Speaker Localization task. This configuration achieves a 1.2\% error rate, significantly better than the base model (3.0\%), indicating that the Visual Delta Encoder alone already exhibits strong visual awareness of speaker location. Subsequently, with both the Visual Base and Visual Delta Encoder features retained, HumanOmni-Speaker further reduces the error rate to 0.8\%. 

% On the more complex VR-SDR task, HumanOmni-Speaker with the Visual Delta Encoder reduces the  \enquote{Who said what}  error rate from 52.1\% to 47.1\%, and simultaneously lowers the \enquote{Who said when} error  rate from 31.5\% to 28.5\%. This indicates that the Visual Delta Encoder, by enhancing visual awareness of speaker location, acts as an effective semantic bridge across modalities, aligning high-frequency audio with sparse visual information and enabling deeper binding of \enquote{Who said what and when}. 

\noindent\textbf{End-to-End vs.\ Alternative Pipelines.}
We compare our end-to-end architecture against two common alternative paradigms. First, a pipeline-based approach that applies Whisper-diarization~\cite{WhisperDiarization} for audio segmentation followed by Qwen3-VL~\cite{Qwen3VL} for visual speaker identification yields significantly higher error rates (64.2\% SA-WER and 39.3\% IER), underscoring the advantage of tightly coupling audio-visual semantics within a single model. Second, replacing raw full-frame input with face-cropped regions from an off-the-shelf detector causes the SI-Hard error rate to rise sharply from 21\% to 38\%. This degradation stems from cascaded detection errors in multi-person scenes and the loss of global contextual cues (head pose, relative position, gesture) that are critical for cross-modal disambiguation. Together, these comparisons validate our end-to-end design, where the Visual Delta Encoder attends to the active speaker's mouth while retaining surrounding scene context for robust identity resolution.

% For comparison, we use pipeline-based approach that first applies Whisper-diarization\cite{WhisperDiarization}  on audio and then uses Qwen3-VL\cite{Qwen3VL} for speaker identification within segmented video clips. This method yields significantly higher error rates (64.2\% for \enquote{Who said what} and 39.3\% for \enquote{Who said when}), underscoring the advantage of our end-to-end architecture in tightly coupling audio-visual semantics. 

\begin{figure}[htbp]
\centering
\includegraphics[width=0.84 \linewidth]{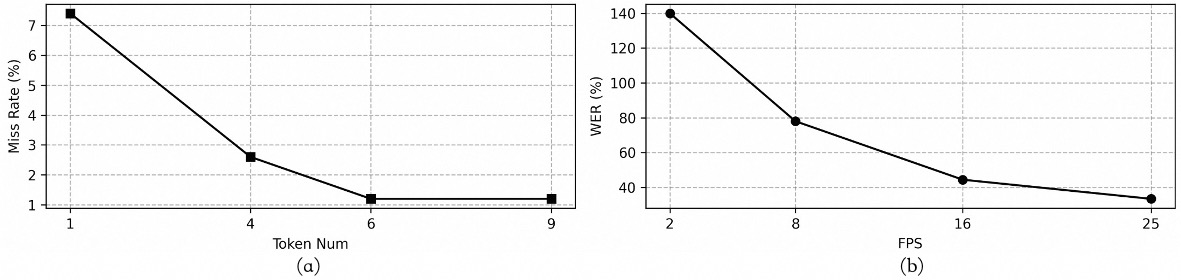}
\vspace{-0.2cm}
\caption{\label{fig:Ablation}(a) Effect of token number on SL task. (b) Effect of  FPS on VSR task.
}
\end{figure}

\noindent\textbf{Hyperparameter Sensitivity.}
We further analyze two key design parameters of the Visual Delta Encoder: the number of structured tokens per frame and the input frame rate. As shown in Fig.~\ref{fig:Ablation}(a), reducing the token count below 6 leads to significant degradation in speaker localization accuracy due to insufficient spatial representation, while increasing it beyond 6 yields minimal gains---indicating that 6 tokens per frame strikes an optimal balance. Fig.~\ref{fig:Ablation}(b) shows that an excessively low frame rate (e.g., 2 FPS) prevents effective lip reading; 16 FPS enables basic capability, while 25 FPS achieves optimal performance for fine-grained viseme modeling.

% To further analyze the design choices of the Visual Delta Encoder, we conduct ablation studies on two key hyperparameters: the number of structured tokens per frame and the input video frame rate (FPS). We analyze the former on the Speaker Localization task and the latter on the VSR task, both of which can be performed using only the Visual Delta Encoder to isolate the effect of each parameter.

% As shown in Fig.\ref{fig:Ablation}(a), reducing the token count per frame below 6 leads to a significant degradation in speaker localization accuracy, due to insufficient representation of spatial displacement.With 6 tokens per frame, the error rate is already 1.2\%,and increasing the token count yields minimal gains, indicating that 6 tokens are sufficient for effective motion modeling. Fig.\ref{fig:Ablation}(b) shows that an excessively low frame rate (e.g., 2 FPS) prevents effective lip reading. While 16 FPS enables basic lip-reading capability, the model achieves optimal performance at 25 FPS, indicating that higher frame rates enhance fine-grained viseme modeling.

\noindent\textbf{Computational Cost.}
Building on the above hyperparameter choices (6 tokens, 25 fps), we measure the resulting overhead on a 5.7\,s clip ($420\times756$, Qwen2.5-Omni-3B, V100). Incorporating Visual Delta tokens increases the token count from $2{,}257$ to $3{,}111$, inference time from $2.62$\,s to $3.13$\,s, peak GPU memory from $13.2$\,GB to $17.2$\,GB, and FLOPs from $26.9$\,T to $38.2$\,T. Compared to a naive 25\,fps ViT approach that incurs $>$10$\times$ token blow-up, our structured design caps the overhead at only $+38\%$ additional tokens, demonstrating a favorable trade-off between high-frequency temporal perception and computational efficiency.

\begin{table}[tbp]
\vspace{0.5em}
\small 
\caption{Ablation Study. Percentages in parentheses denote the relative improvement compared to the baseline (visual base only). }
\label{tab:ablation_study}
\vspace{-0.2cm}
\centering
\resizebox{0.8\textwidth}{!}{ 
\begin{tabular}{>{\centering\arraybackslash}p{0.35\linewidth}cccc}
\toprule
  \textbf{}&\makecell{\textbf{Feature-type} \\ \textit{visual base} | \textit{visual delta} | \textit{audio}} & \makecell{\textbf{Speaker} \\ \textbf{Loc. } ($\downarrow$)} & \makecell{\textbf{VR-SDR} \\ \textbf{What} ($\downarrow$)} & \makecell{\textbf{VR-SDR} \\ \textbf{When} ($\downarrow$)} \\ 
\midrule
 \textbf{pipeline-based}&Yes | No | Yes & -& 64.2& 39.3\\
 \midrule
 \multirow{3}{*}{\centering \textbf{HumanOmni Speaker}}& Yes | No | Yes & 3.0 & 52.1&31.5\\
 &No | Yes | Yes & 1.2 & - & - \\ 
 &Yes | Yes | Yes& \textbf{0.8} \scriptsize{(-73\%)}& \textbf{47.1} \scriptsize{(-9.6\%)}& \textbf{28.5} \scriptsize{(-9.5\%)}\\ 
\bottomrule

\end{tabular}
}
\end{table}

\section{Conclusion}

In this work, we presented HumanOmni-Speaker, a unified Omni-modal LLM specifically designed for complex, speaker-centric interactions. By introducing a high-frame-rate Visual Delta Encoder, we addressed the architectural perception gap, facilitating genuine end-to-end joint modeling of \enquote{\textit{Who said what and when}.} Furthermore, to systematically mitigate visual shortcuts and overcome current evaluation limitations, we proposed the rigorous VR-SDR paradigm and the comprehensive HumanOmni-Speaker Benchmark. Extensive experiments demonstrate that our model significantly outperforms existing open-source baselines across a diverse spectrum of tasks—ranging from fine-grained lip-reading and spatial localization to holistic speaker diarization and recognition.

\clearpage
% ---- Bibliography ----
% Bibliography inlined from main.bbl for arXiv upload (no BibTeX run required).
% Original setup (kept for reference):
% \bibliographystyle{splncs04}
% \bibliography{main}

\end{document}